\pdfoutput=1

\documentclass[11pt]{article}

\usepackage[]{naacl2021}

\usepackage{times}
\usepackage{latexsym}

\usepackage[T1]{fontenc}

\usepackage[utf8]{inputenc}

\usepackage{microtype}
\usepackage{amsmath,amsfonts,amsthm,amssymb}
\usepackage{xspace}
\usepackage{bbm}
\usepackage{booktabs}
\usepackage{graphicx}
\usepackage{microtype}
\usepackage[normalem]{ulem}
\usepackage{array}
\usepackage{calc}
\usepackage{pifont}
\usepackage{makecell}
\definecolor{darkgreen}{RGB}{0,140,20}

\newcommand\ML{M\&L\xspace}
\newcommand\lp{\ensuremath{\ell_+}}
\newcommand\lm{\ensuremath{\ell_-}}
\newcommand{\lemmas}{\ensuremath{\mathcal{L}}\xspace}
\newcommand{\indicator}[1]{\ensuremath{\mathbbm{1}\left[#1\right]}}
\newcommand{\PMc}{\ensuremath{P_{M}}}
\newcommand{\mw}[1]{{\color{violet} #1}}
\newcolumntype{P}[1]{>{\centering\arraybackslash}p{#1}}
\newcolumntype{M}[1]{>{\centering\arraybackslash}m{#1}}
\newcommand{\AnD}{\hskip 2em plus 1fil minus 0.5em}

\title{Refining Targeted Syntactic Evaluation of Language Models}

\author{Benjamin Newman \AnD Kai-Siang Ang \AnD Julia Gong \AnD John Hewitt \\
Department of Computer Science\\
Stanford University \\
\{\texttt{blnewman,kaiang,jxgong,johnhew}\}\texttt{@cs.stanford.edu}}

\begin{document}
\maketitle
\begin{abstract}
Targeted syntactic evaluation of subject-verb number agreement in English (TSE) evaluates language models' %
syntactic knowledge using hand-crafted minimal pairs of sentences %
that differ only in the main verb's conjugation.
The method evaluates whether language models rate each grammatical sentence as more likely than its ungrammatical counterpart. %
We identify two distinct goals for TSE\@.
First, evaluating the \textit{systematicity} of a language model's syntactic knowledge: given a sentence, can it conjugate arbitrary verbs correctly?
Second, evaluating a model's \textit{likely behavior}: given a sentence, does the model concentrate its probability mass on correctly conjugated verbs, even if only on a subset of the possible verbs?
We argue that current implementations of TSE do not directly capture either of these goals, and propose new metrics to capture each goal separately.
Under our metrics, we find that TSE overestimates systematicity of language models, but that models score up to 40\% better on verbs that they predict are likely in context.

\end{abstract}

\section{Introduction}

As neural language models have emerged as both broadly useful engineering tools~\cite{devlin2018bert, Radford2019LanguageMA} and potential models of human language processing~\cite{linzen2018distinct, ettinger2018assessing, futrell2019neural}, evaluations targeting their syntactic ability have been developed to better understand their capabilities.

One such method for syntactic evaluation tests models' knowledge of English subject-verb (S/V) number agreement \citep{linzen2016assessing,gulordava2018colorless}.
These studies consider minimal pairs of sentences, such as \textit{The keys to the cabinet is/are on the table}, that differ only in verb number, and %
test if models rate grammatical sentences as more probable.
The syntactically correct of the two sentences is sampled from natural corpora \citep{linzen2016assessing,kuncoro2018lstms} or constructed from templates.
The use of templates, known as Targeted Syntactic Evaluation (TSE), allows for the fine-grained evaluation of models on specific, often rare, syntactic phenomena~\cite{marvin2018targeted, ettinger2018assessing, warstadt2019blimp}, but (when evaluating S/V number agreement) relies on researchers hand-specifying a small set of verb lemmas that are substituted into each template.

In this work, we improve the TSE methodology by disentangling its broad objective of evaluating syntactic ability into two distinct goals, and we introduce two variants of TSE to separately capture each goal.
These evaluations demonstrate that neural models do not generally consider well-conjugated verbs more likely than their incorrect conjugations, but instead prefer to correctly conjugate verbs they deem likely.

We argue that the objective of evaluating syntactic ability can be decomposed into two goals and that current implementations of TSE do not achieve either of them. %
The first goal is measuring \textbf{systematicity}: for a specific syntactic construction, does the model correctly conjugate arbitrary verbs with the grammatical number of the subject?
TSE currently fails to capture this because it evaluates models using only a small set of verbs for each syntactic construction.
If models only conjugate these verbs correctly, they receive a high score, even if they conjugate other verbs incorrectly.
The second goal is measuring \textbf{likely behavior}: when we sample verbs from the model in a specific syntactic construction, will they be properly conjugated?
TSE fails to directly capture this because the small set of verbs used in evaluation might differ from the verbs that are likely in context under the model.
If models conjugate these hand-specified verbs incorrectly, they receive a low score, even if they correctly conjugate more likely verbs.

\begin{table}[]
    \centering
    \includegraphics[width=\linewidth]{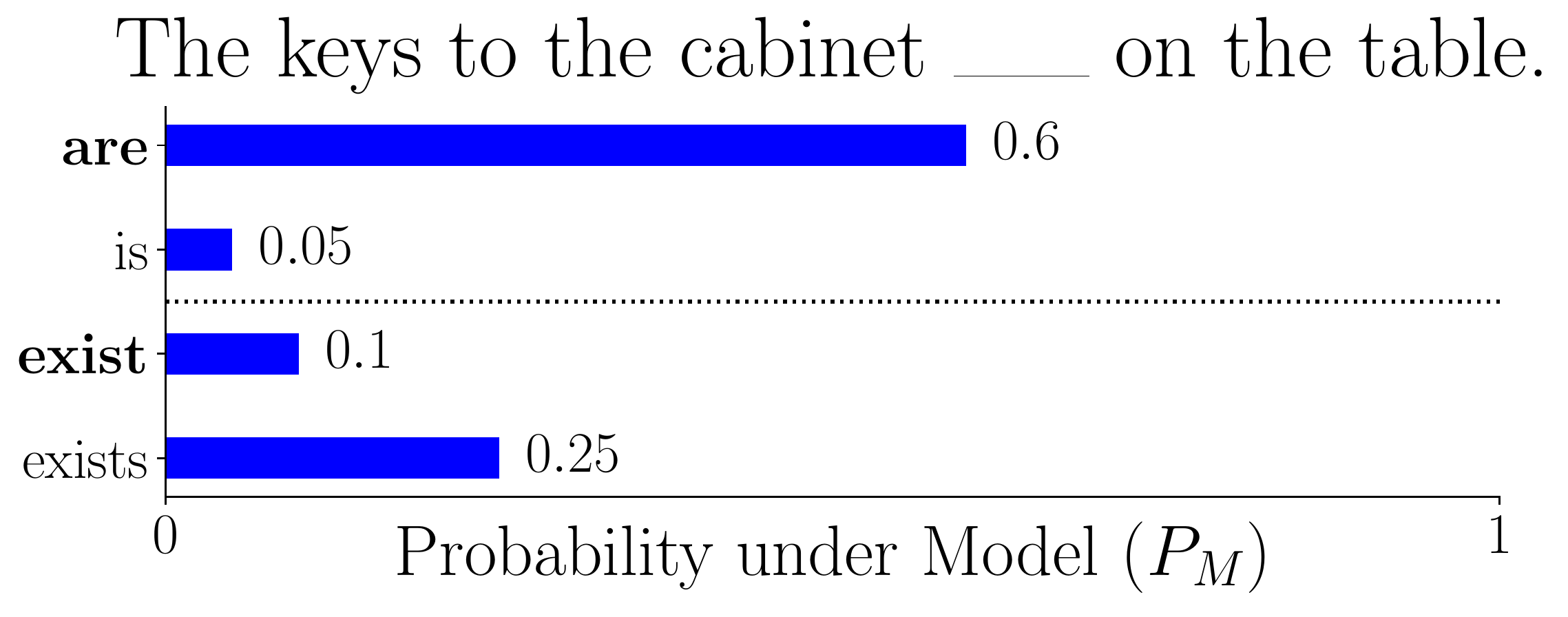}
    \begin{tabular}{M{20mm} M{35mm}  M{10mm}}
     \toprule
         Metric & Computation & Score \\
         \midrule
         TSE & \textbf{are} $>$ is %
         & $1.0$ \\
         \midrule
         \shortstack{EW \\ {\small(systematicity)}} 
         & \shortstack{\textbf{are} $>$ is %
         \\ exists $>$ \textbf{exist}} %
         & $0.5$\\
         \midrule
         \shortstack{MW \\\;\\{\small(likely behavior)}}& 
         $\dfrac{\textbf{are} + \textbf{exist}}{\textbf{are} + \textbf{exist} + \text{is} + \text{exists}}$ & $0.7$ \\
         \bottomrule
    \end{tabular}
    \caption{A toy example where a language model puts more probability mass on the correct \textit{are} and the incorrect \textit{exists},
    showing how TSE currently may not fully capture a model's syntactic ability.
    In contrast, we propose EW, which captures this failure of systematicity, and MW, which captures the probability of sampling a correct conjugation. Bolded verbs are correct.}  
    \label{tab:toy_example}
\end{table}

To motivate these goals and the misspecification of TSE, consider evaluating a language model on the following two pairs of sentences:
\begin{center}
    \textit{The keys to the cabinet is/are on the table.}\\
    \textit{The keys to the cabinet exist/exists on the table.}\\
\end{center}
\noindent where for simplicity we assert that the only possible verbs are: is/are (\texttt{be}) and exists/exist (\texttt{exist}). 
Let the model assign higher probability mass to the correct conjugation for the \texttt{be} pair but not for the \texttt{exist} pair (Table~\ref{tab:toy_example}).

First, consider evaluating systematicity.
To reflect how TSE chooses a small subset of the possible verbs for evaluation, in this toy example let it choose only \texttt{be}.
Thus, the model scores $1$ out of $1$, whereas a test of systematicity should penalize the model for incorrectly conjugating \texttt{exist}.

Now, consider evaluating likely behavior.
Let this same model generate either of the two correct conjugations (are/exist) with total probability of $0.7$ and generate either of the incorrect conjugations with total probability $0.3$.
Thus, when we sample from the model, it generates a correct conjugation with probability $0.7$, but TSE cannot measure this, since it gives a binary score to each verb pair. %

The first of our proposed evaluations, \textbf{equally-weighted syntactic evaluation} (EW), addresses systematicity.
To better approximate a model's ability to conjugate \textit{any} verb, EW expands TSE to consider a much larger set of verbs than given in the templates used by prior work.

The second of our proposed evaluations, \textbf{model-weighted syntactic evaluation} (MW), addresses likely behavior.
This method computes the probability mass that models put on producing the correct verb conjugation given a particular syntactic context.
It rates the syntactic quality of samples---models need not conjugate \textit{all} verbs, but instead be likely to generate \textit{some} well-conjugated verb.

We conduct these evaluations on four pretrained language models using two template datasets: \ML~\cite{marvin2018targeted} and BLiMP~\cite{warstadt2019blimp}.
Overall, we find that the EW scores are lower than the TSE scores, indicating that the verb choices in these templates overestimate models' systematicity with respect to subject-verb number agreement.
This lack of systematicity is particularly apparent when we test verb lemmas that models find unlikely, with scores dropping by up to $40\%$. %
In contrast, the MW scores are high, suggesting that language models preferentially conjugate verbs they deem likely.
Moreover, this ability improves when the tail of the distribution is truncated, as it is in decoding strategies like nucleus sampling~\cite{Holtzman2020The}.\footnote{Code available at \url{https://github.com/bnewm0609/refining-tse}}

\section{Methods}

To define our metrics, we introduce some notation. TSE has two components: the model $M$ to evaluate, and the set of templates $T$ with interesting syntactic phenomena (e.g., from \citet{marvin2018targeted}).
In S/V number agreement, each template contains a context $c$, including the subject that specifies the correct verb inflection; and the verb lemma $\ell$ with correct and incorrect inflections in the third person present tense (\lp{} and \lm, respectively).
$M$ takes $c$ and produces a distribution $\PMc(\cdot \mid c)$ over its vocabulary, which we assume includes \lp{} and \lm.
We then compute a score for each template and average the scores over all templates to get a final score for $M$.
The TSE score for a template can be expressed as:
\begin{align}
\label{eq:tse}
    \indicator{\PMc(\lp \mid c) > \PMc(\lm \mid c)}.
\end{align}

The crux of our proposal is to use a large set  of lemmas, \lemmas, %
while drawing contexts $c$ from a predefined set of templates $T$.
We define two evaluation methods using \lemmas:
\paragraph{Equally-Weighted (EW)} Here we average (\ref{eq:tse}) over all $\ell$ in \lemmas, evaluating systematicity.

\paragraph{Model-Weighted (MW)} Here we compute the total probability of generating a correct inflection conditioned on generating a lemma in \lemmas: 
\begin{align}
    \frac{\sum_{\ell \in \lemmas}\PMc(\ell_+ \mid c)}{\sum_{\ell \in \lemmas}\PMc(\ell_+ \mid c) + \PMc(\ell_- \mid c)},
\end{align}
evaluating likely behavior.
See Table~\ref{tab:toy_example} for how these are computed in the toy example.

\section{Experiments}
\paragraph{Data}
We use S/V number agreement TSE templates from \citet{marvin2018targeted} and BLiMP \citep{warstadt2019blimp} (for BLiMP we use the minimal pairs differing in verb, not subject). For our MW and EW evaluations, we only keep templates with unique contexts (i.e., templates not differing solely in verb lemma).
We also ensure that all sentences start with a capital letter (for cased models) and end with a sentence-final period (for bidirectional models).

Our list of English verb lemmas contains 3,562 lemmas and is compiled by %
combining the top 1,000 most frequent verb lemmas from COCA, extracting all tokens with the \texttt{VB} part-of-speech tag in the Penn Treebank (1,951 lemmas), and scraping 3,250 lemmas from the Giant Verb List \citep{davies2008corpus, marcus1993building, giantVerbList}.
\footnote{The verb lemmas are accessible from the Appendix.}
Masked LMs may assign a different number of tokens to plural and singular forms of the same lemma, and they may not model joint probabilities over multiple tokens.
To enable a fairer comparison between LMs and masked LMs, we only consider lemmas where both inflections are in the wordpiece vocabulary of the models.
This choice leaves 980 lemmas for BERT cased, 1159 for BERT uncased, and 1265 for GPT2 and RoBERTa (so results are not comparable between models).
This verbal variety situates our work between \citet{gulordava2018colorless}'s and \citet{marvin2018targeted}'s:
our verbs can be infelicitous like the sentences in \citet{gulordava2018colorless}, but our contexts are felicitous. See Section~\ref{sec:related} for additional discussion.

 \paragraph{Models}
We evaluate both bidirectional and unidirectional models, including BERT-large-uncased, BERT-large-cased, GPT2-XL, and RoBERTa-large~\cite{devlin2018bert, Radford2019LanguageMA, liu2019roberta}, all using the Huggingface Tranformers library~\cite{wolf-etal-2020-transformers}.

To understand models' performances at the head and tail of their distributions, we additionally restrict \lemmas{} to the lemmas assigned high and low probabilities.

To consider the high-confidence lemmas, for each template in the dataset, we record the MW and EW scores computed using the inflections that fall into the top $p$ percentile of the model's distribution.
We choose $p \in \{10, 20, 30, 40, 50, 60, 70, 80, 90, 95, 97, 100\}$, noting that for each $p$, the distribution we use is the same as the one used by nucleus sampling (with a nucleus of size $p$).

Analogously, to focus on the low-confidence lemmas, we consider the lemmas where both inflections fall into the bottom $p$ percentile of the model's distribution. %
Here, we choose $p \in \{50, 10, 1, 0.1, 0.01, 0.001, 0.0001\}$.\footnote{At times, a cut-off lies within the probability mass on an inflection of interest.
In these cases, we linearly interpolate between scores with and without the inflection included.}

\section{Results}

\begin{table*}
    \centering
    \small
    \begin{tabular}{c | ccc | ccc | ccc | ccc}
\toprule
Templates & \multicolumn{3}{c|}{BERT cased} & \multicolumn{3}{c|}{BERT uncased} & \multicolumn{3}{c|}{RoBERTa} & \multicolumn{3}{c}{GPT2} \\
 & \mw{MW} & EW & TSE & \mw{MW} & EW & TSE & \mw{MW} & EW & TSE & \mw{MW} & EW & TSE \\
\midrule
Simple & \mw{0.99}& 0.94 & \textbf{1.00} & \mw{0.98}& 0.90 & \textbf{1.00} & \mw{0.98}& 0.93 & \textbf{1.00} & \mw{0.90}& 0.86 & \textbf{1.00} \\
In a sentential complement & \mw{0.92}& 0.67 & \textbf{0.89} & \mw{0.92}& 0.60 & \textbf{0.86} & \mw{0.92}& 0.67 & \textbf{0.88} & \mw{0.96}& 0.65 & \textbf{0.89} \\
VP coordination & \mw{0.91}& 0.89 & \textbf{0.90} & \mw{0.93}& \textbf{0.90} & \textbf{0.90} & \mw{0.93}& 0.90 & \textbf{0.93} & \mw{0.89}& 0.87 & \textbf{0.97} \\
Across prepositional phrase & \mw{0.91}& 0.83 & \textbf{0.93} & \mw{0.83}& 0.75 & \textbf{0.85} & \mw{0.87}& 0.83 & \textbf{0.89} & \mw{0.84}& 0.76 & \textbf{0.96} \\
Across subject relative clause & \mw{0.87}& \textbf{0.84} & \textbf{0.84} & \mw{0.88}& 0.84 & \textbf{0.85} & \mw{0.76}& 0.72 & \textbf{0.80} & \mw{0.82}& 0.77 & \textbf{0.97} \\
Across object relative clause & \mw{0.91}& 0.88 & \textbf{0.91} & \mw{0.86}& 0.80 & \textbf{0.85} & \mw{0.88}& 0.85 & \textbf{0.91} & \mw{0.95}& 0.89 & \textbf{0.99} \\
Across object relative (no that) & \mw{0.92}& 0.88 & \textbf{0.90} & \mw{0.79}& 0.72 & \textbf{0.81} & \mw{0.86}& 0.82 & \textbf{0.89} & \mw{0.95}& 0.89 & \textbf{0.99} \\
In object relative clause & \mw{0.93}& 0.95 & \textbf{0.97} & \mw{0.95}& 0.97 & \textbf{0.99} & \mw{0.89}& 0.91 & \textbf{0.97} & \mw{0.91}& 0.88 & \textbf{0.98} \\
In object relative (no that) & \mw{0.90}& 0.91 & \textbf{0.92} & \mw{0.81}& \textbf{0.82} & \textbf{0.82} & \mw{0.82}& 0.83 & \textbf{0.90} & \mw{0.91}& 0.88 & \textbf{0.97} \\
BLiMP & \mw{0.81}& 0.73 & \textbf{0.90} & \mw{0.78}& 0.69 & \textbf{0.85} & \mw{0.70}& 0.66 & \textbf{0.78} & \mw{0.82}& 0.75 & \textbf{0.91} \\
\bottomrule

    \end{tabular}
        \caption{MW, EW, and TSE evaluations on various models and syntactic constructions (See \citet{warstadt2019blimp, marvin2018targeted} for descriptions). MW is colored differently because its score is based directly on the model's probability mass, while EW and TSE are based on 0/1 judgements, so they are not directly comparable.
    }
    \label{tab:exp_results}
\end{table*}

\begin{figure}
    \centering
    \includegraphics[width=.9\linewidth]{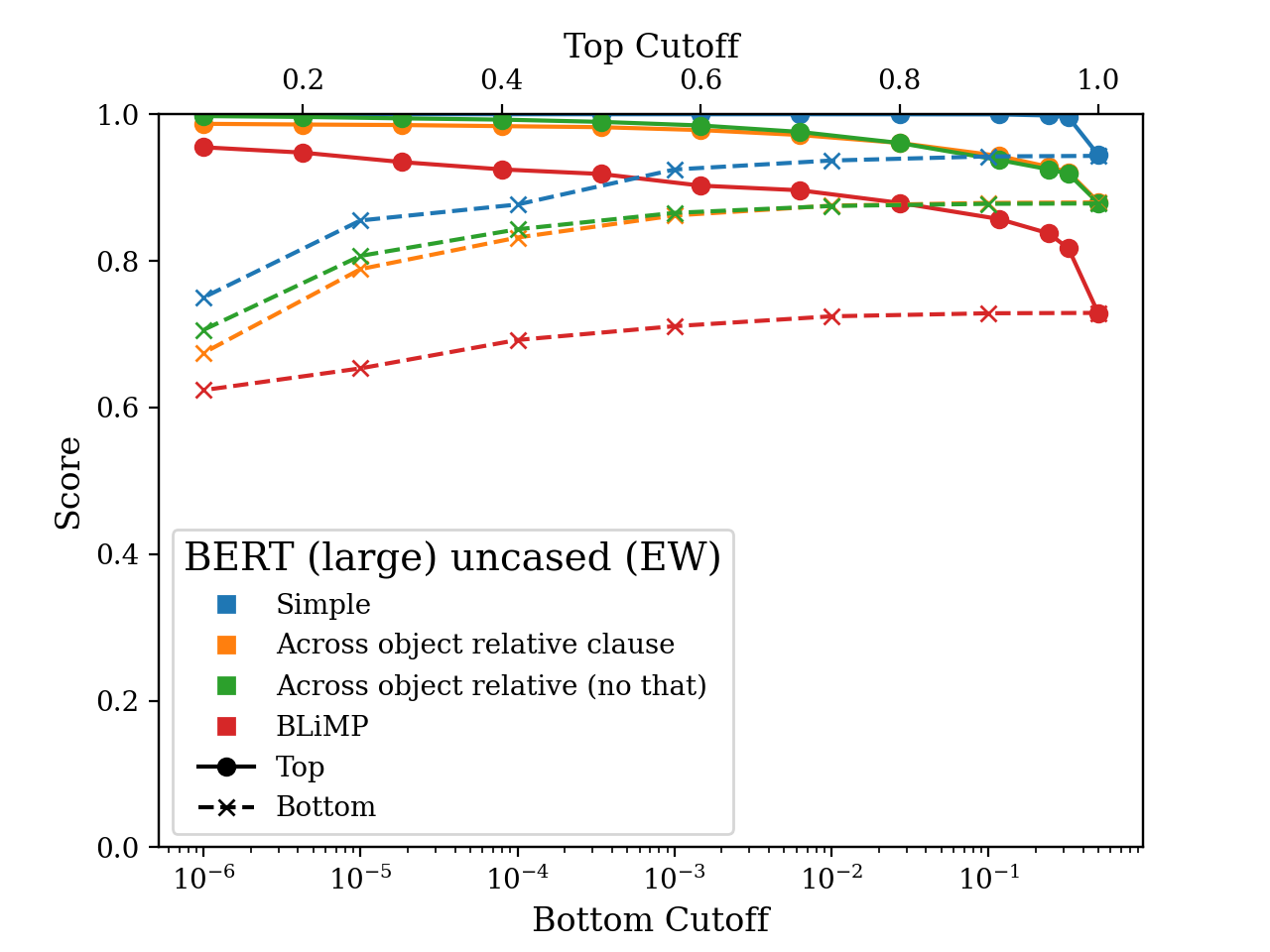}
    \includegraphics[width=.9\linewidth]{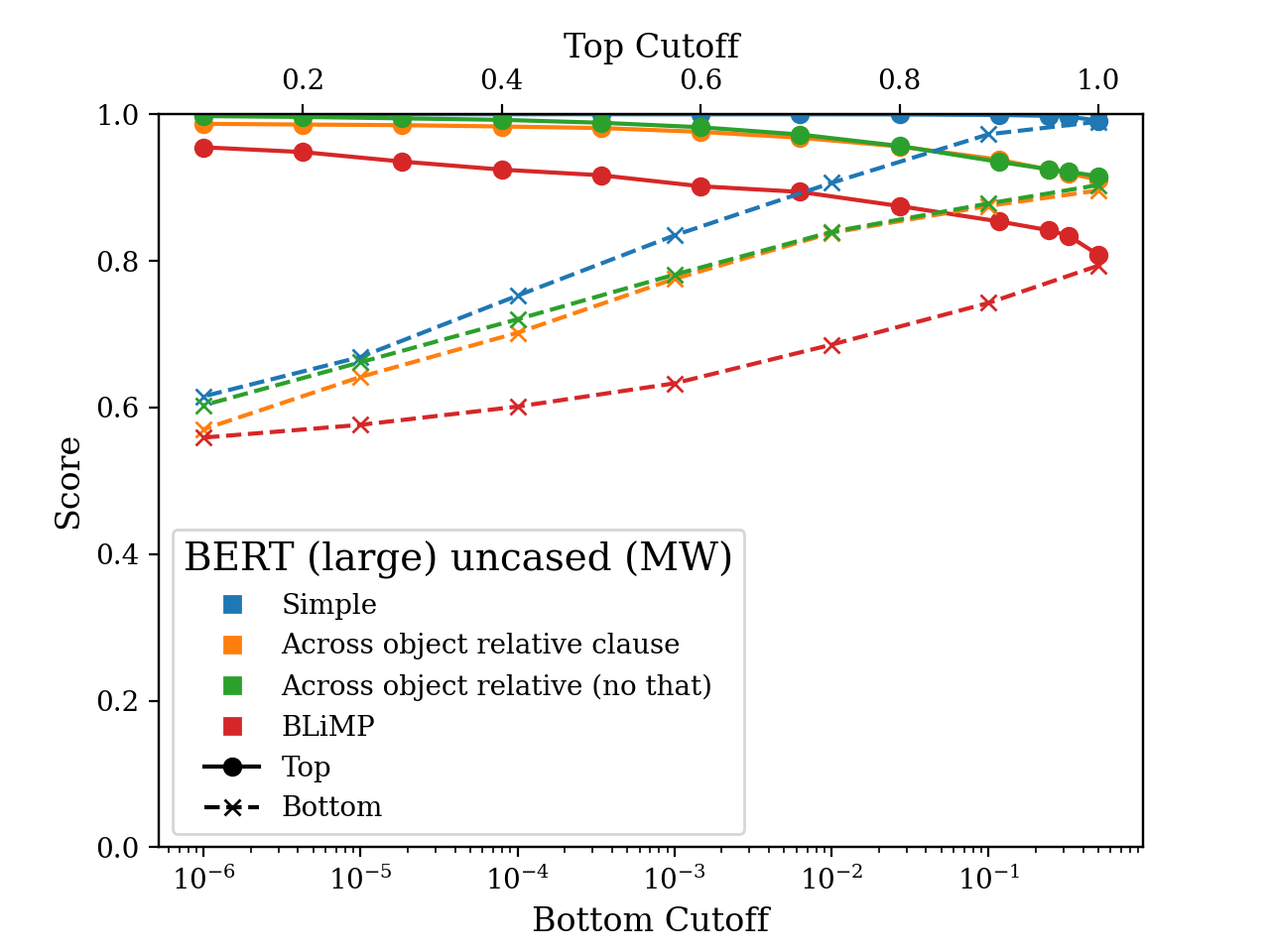}
    \caption{EW and MW scores as a function of Top $p$ and Bottom $p$ cutoffs for a subset of syntactic constructions (colors) using the BERT cased model.}%
    \label{fig:top-bottom-k}
\end{figure}

Our results can be found in Table~\ref{tab:exp_results}.
We find that EW scores are almost always lower than TSE scores, indicating that TSE overestimates systematicity.
On the other hand, higher MW scores reveal that sampling from the models is likely to result in correct conjugations.

A potential confounder for unidirectional LMs (GPT2) is that they only receive the left context and subject verb pairs sometimes look like noun phrases.
For example, a sentence starting with \textit{The officer} can be continued by \textit{experiences joy} or by \textit{experience is overwhelming}. This is not an issue when there are phrases or clauses between the subject and verb, and it may not occur for other English syntactic phenomena or in other languages.

To investigate the extent to which models perform well on likely lemmas and poorly on unlikely lemmas, we plot these scores for the top and bottom $p$ percentiles in Figure~\ref{fig:top-bottom-k}. %
In general, the models perform better on lemmas that they assign high probability to in both evaluations.

For example, consider the BERT cased model assessed on object relative clause constructions. The MW plot illustrates that sampling from the top $60\%$ of the distribution will produce a grammatical output with $97\%$ probability, while sampling from the entire distribution only does so with $91\%$ probability.
The EW plot shows that the model attains a score under $80\%$ when assessed on verbs in the bottom $0.001\%$ of the model's probability mass, even though considering verbs in the top $90\%$ of the model's probability mass would yield a score over $94\%$.
These observations extend previous work on nucleus sampling, showing that cutting off the tails of the distribution generates more syntactically correct outputs~\cite{Holtzman2020The}. %

There are two additional factors to keep in mind for these plots.
First, the heads and tails of the distributions often contain very few lemmas eligible for use in score calculation.
Second, models often assign probability mass to other lemma inflections (e.g.\ the past tense) that do not allow us to assess models' S/V number agreement ability.
See the Appendix for related plots.

\subsection{Qualitative Results}
Earlier, we motivated MW with the consideration that the lemmas TSE uses might be unlikely, and therefore give an unrealistic depiction of models' likely syntactic behavior.
Two examples where this happens and leads to a deceptively low score on a template for a model (here BERT-large-cased) are in Table~\ref{tab:qualitative_examples}.

In the first column, the lemma set used by TSE contains \texttt{like}, \texttt{hate}, and \texttt{love}, and the model puts more probability on \textit{like} than \textit{likes}, leading to a TSE score of $0.67$.
However, the most probable lemmas are \texttt{meet}, \texttt{encounter}, \texttt{see}, and \texttt{face}, all of which the model conjugates correctly.

In the second column, there is another example where the MW score rewards models for correct conjugations while TSE does not.
Like the last example, the lemma set used by TSE contains \texttt{like}, \texttt{hate}, and \texttt{love}, and \texttt{like} is conjugated incorrectly.
However, the more probable lemmas \texttt{pilot}, \texttt{control}, \texttt{employ}, \texttt{train}, \texttt{use}, \texttt{include}, \texttt{have}, \texttt{order}, \texttt{command}, and \texttt{feature} are all conjugated correctly.
\begin{table}[]
    \centering
    \begin{tabular}{rl | rl }
     \toprule
     \multicolumn{2}{P{35mm}}{The senators that the skater [mask] are young.} &  \multicolumn{2}{P{35mm}}{
     The pilots that the executive [mask] are tall.}\\
     \midrule
        meets & 0.20       &%
        pilots & 0.088 \\
        
        encounters & 0.057 &%
        controls & 0.059 \\
        
        sees & 0.057       &%
        employs & 0.025 \\
        
        meet & 0.048       &%
        trains & 0.023 \\
        
        encounter & 0.023  &%
        uses& 0.022 \\
        
        see & 0.019        &%
        includes & 0.019 \\
        
        \#\#s & 0.018      &%
        has &  0.017 \\
        
        faces & 0.013      &%
        orders & 0.015 \\
        
        saw & 0.012        &%
        commands & 0.014 \\
        
        met & 0.010        &%
        features & 0.013 \\
    \bottomrule
    \end{tabular}
    \caption{Example sentences and the top 10 most probable subwords.}
    \label{tab:qualitative_examples}
\end{table}

\section{Related Work}\label{sec:related}

\paragraph{Evaluating Models} Some previous work has focused on using minimal pairs to evaluate syntactic representations of models.
\citet{goldberg2019assessing} and \citet{wolf2019some} assess the syntactic abilities of large transformers like BERT and GPT2, while \citet{kuncoro2018lstms}, \citet{tran2018importance} and \citet{kim2019unsupervised} evaluate architectures designed to capture syntax (e.g., Ordered Neurons~\cite{shen2018ordered} and Recurrent Neural Network Grammars~\cite{dyer2016recurrent}).
In these cases, minimal pair evaluations should align with models' performance as language models, which is measured by our MW score.

\paragraph{Psycholinguistics} Recent work has also applied experimental procedures from psycholinguistics to compare human and neural model language processing \citep{futrell2019neural}.
Experiments investigating garden path sentences' surprisal, S/V number agreement, and other specific syntactic phenomena reveal that models and humans have different patterns of errors and processing \citep{linzen2018distinct, ettinger2018assessing, wilcox-etal:2020-on-the-predictive-power, van2020single}.
Many of these phenomena are rare, so evaluations with templated minimal pairs complement perplexity as a metric for evaluating models’ syntactic generalization \citep{hu2020systematic}.
When measuring syntactic ability on arbitrary lemmas, our EW metric would be preferred.

\paragraph{Lexical Choice in Syntactic Evaluation}
Prior work also considered how the lexical items in minimal pairs affect the syntactic evaluation of models.
\citet{marvin2018targeted} note that certain verbs are preferentially conjugated correctly (they observe RNNs conjugate \texttt{be} correctly more often than \texttt{swim}) and claim that this is due to unigram frequency of the verbs.
Similarly, we observe that models succeed on our MW metric indicating that they correctly inflect verbs with high in-context probability under the model.

Relatedly, \citet{yu-etal-2020-word} investigate the nouns used in TSE minimal pairs and find that language model performance at subject-verb number agreement is uncorrelated with unigram probability of the noun. 
We instead focus on model-estimated in-context probability of the verb in minimal pairs, finding that model performance increases with the model probability.

Finally, \citet{gulordava2018colorless} argue that the results of syntactic evaluations are influenced by semantic associations between tokens, so they remove these associations by substituting each token with a different randomly selected token with the same syntactic role.
Although the resulting minimal pairs are infelicitous, models are still able to predict the correct inflection with above-chance accuracy.
Our methods are similar in that some of the verbs in our evaluation set are infelicitous, however the contexts we use are semantically coherent.
Rather than avoiding semantic effects by creating infelicitous contexts, we marginalize them out by using a large set of verb lemmas.
This makes our metrics less stringent than those of \citet{gulordava2018colorless}, but captures a potentially more realistic setting where we expect our models to perform systematically.

\section{Conclusion}

As neural models have proven successful at NLP tasks and as potential psycholinguistic models, we continue to refine our understanding of how and whether they capture human-like language faculties.
TSE provides a useful framework to address this question, but its current implementation focuses on a limited group of hand-chosen verbs, so it inaccurately reflects models’ syntactic generalization abilities.
In response, we propose two minimal pair evaluations: equally-weighted and model-weighted syntactic evaluation.
The first focuses on \textit{systematicity} by expanding the set of verbs TSE considers, and illustrates that language models still struggle with S/V agreement for unlikely verbs.
The second focuses on \textit{likely behavior} by computing the probability of producing a correctly conjugated verb, and illustrates that despite systematic shortcomings, language models generate syntactically valid utterances with high probability.
By introducing these metrics, we hope to arrive at a clearer picture of the syntactic abilities of language models.

\section{Ethical Considerations}
The metrics we propose have been developed specifically with corpora using Standard American English in order to evaluate models' abilities to understand Standard American English syntax.
This focus means that models performing well under these evaluations may perform poorly in other English dialects, and they may not understand all syntactic systems, for example in other languages.
Finally, our MW metric concerns itself with how models are likely to preform during generative processes (such as beam search or sampling). Performing well on this metric means models will be able to generate more human-like text which has potential downstream harms such as misinformation generation or other inauthentic behavior in situations where written language is the medium used for communication.

\section*{Acknowledgments}
The authors would like to thank the reviewers for their helpful feedback, along with Tal Linzen, Chris Manning, Rishi Bommasani, Kawin Ethayarajh, Lisa Li, Nelson Liu, Yasuhide Miura, Aaron Mueller, and Tianyi Zhang for their invaluable comments and discussions. JH was supported by an NSF Graduate Research Fellowship under grant number DGE- 1656518, and by Two Sigma under their 2020 PhD Fellowship Program.

\bibliography{anthology,naacl2021}
\bibliographystyle{acl_natbib}

\appendix
\section{Additional Plots}
\begin{figure*}
    \centering
    \includegraphics[width=.98\linewidth]{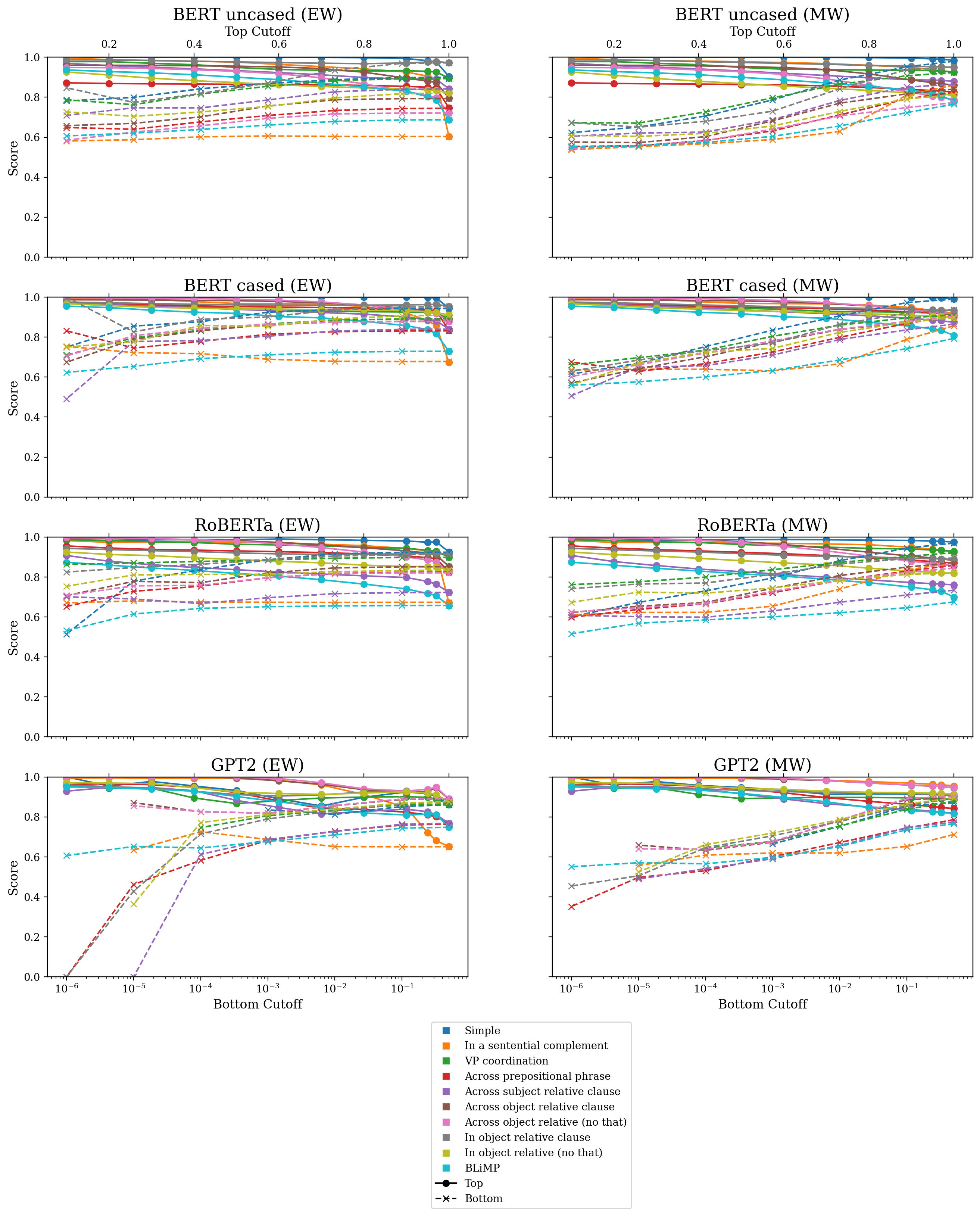}%
        \caption{The plots above show the EW and MW scores as a function of Top-$p$ and Bottom-$p$ percentile cutoffs for BERT-base-uncased, BERT-large-cased, RoBERTa-large and GPT2-XL\@. In general, as the percentile increases, so does the score, though RoBERTa and BERT's EW scores are quite stable.}
    \label{fig:top-bottom-k-scores}
\end{figure*}

\begin{figure*}
    \centering
    \includegraphics[width=.98\linewidth]{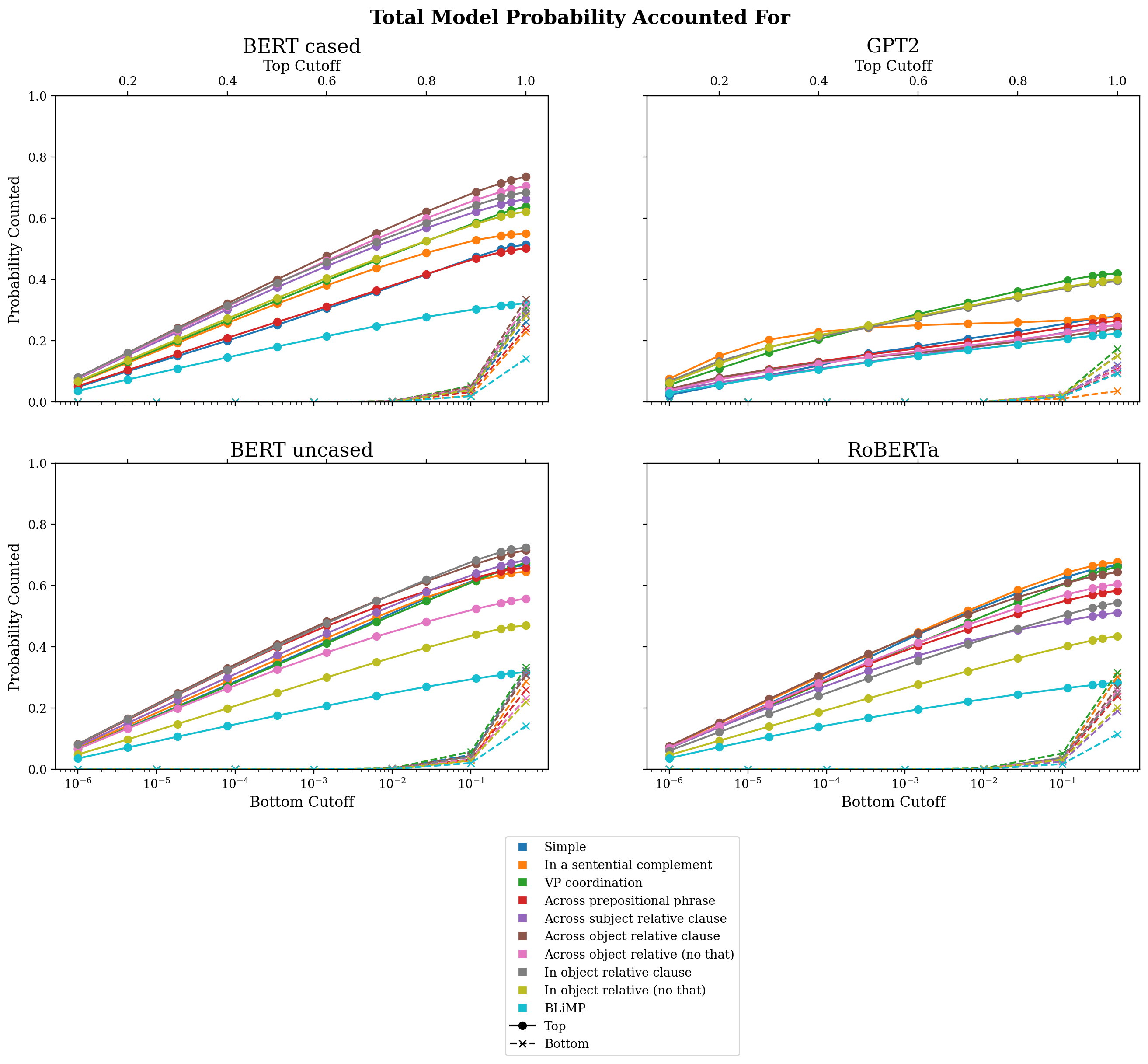}%
        \caption{Above are plots of the proportion of the models' probability mass take up by the inflections of the verb lemmas we used. The y-axis is the total probability mass the lemmas account for and the x-axis is the percentile cutoff. Note that even when considering all of the lemmas, (at $p=100$\%) there is probability mass not covered by our inflections. This probability mass is often put on other inflections of verbs (e.g.\ past-tense verbs) or other vocabulary items.}
    \label{fig:top-bottom-k-probcounted}
\end{figure*}

\begin{figure*}
    \centering
    \includegraphics[width=.98\linewidth]{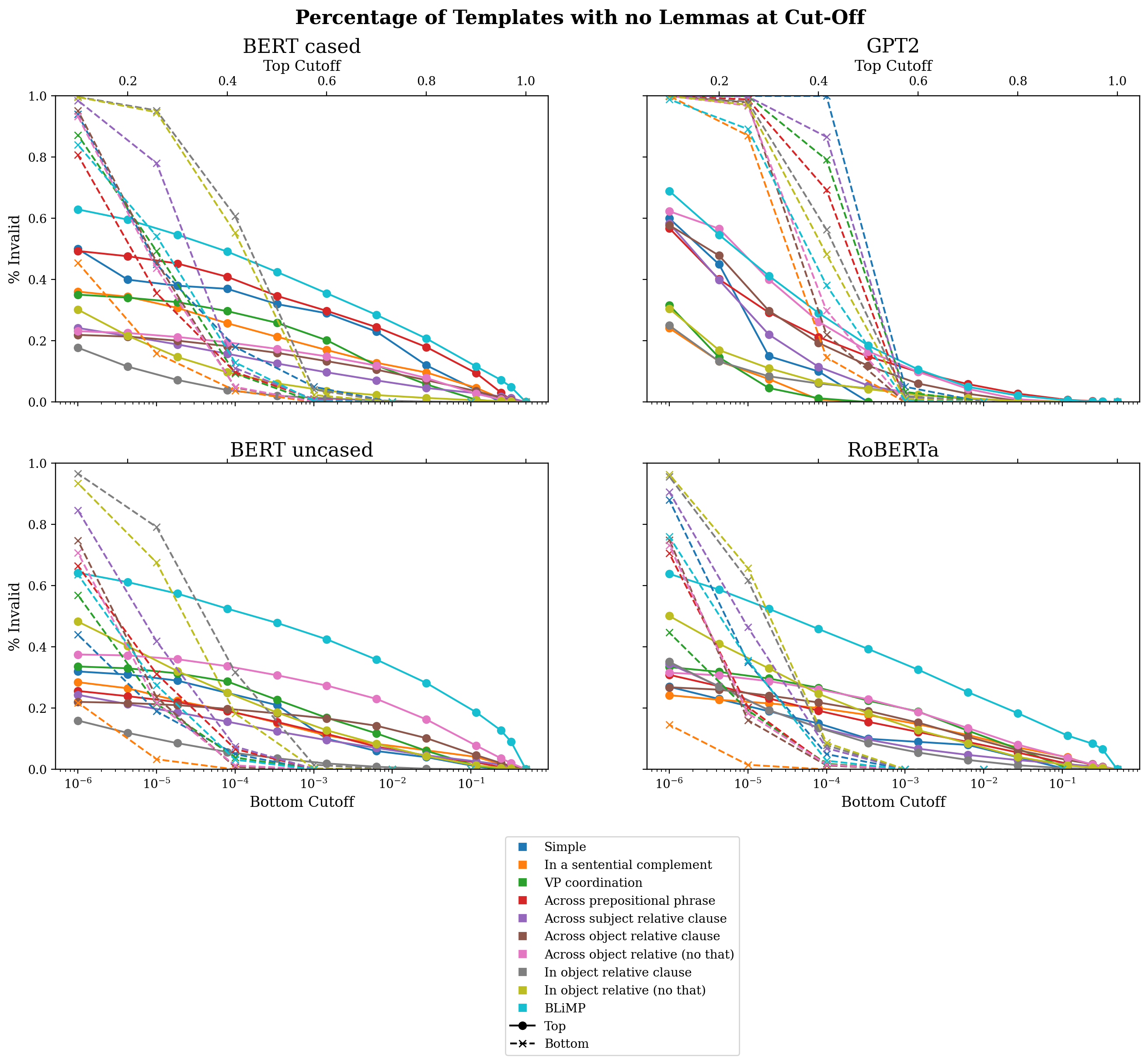}%
        \caption{Above is the proportion of the templates in the datasets where models assign no probability mass to lemmas in the top or bottom $p$\% of the their distributions. The y-axis is the proportion of lemmas that are rejected (i.e.\ values closer to one mean that the scores are calculated based on fewer templates). The x-axis is again the percentile cutoff. Note that the bottom-most cutoffs often has a large proportion of invalid lemmas, so these scores are based on fewer lemmas.}
    \label{fig:top-bottom-k-invalid}
\end{figure*}

\newpage
\section{A Subset of Verb Lemmas}
The $1971$ lemmas from COCA and The Penn Treebank we used are available below~\cite{davies2008corpus, marcus1993building}:

abandon, abate, abide, abolish, absorb, abuse, accede, accelerate, accept, access, acclaim, accommodate,  accomodate, accompany, accomplish, accord, account, accrue, accuse, achieve, acknowledge, acquire, acquit,  act, adapt, add, address, adhere, adjust, administer, admit, adopt, adorn, advance, advantage, advise, affect,  afford, aggravate, agonize, agree, aid, aim, air, alert, alienate, allay, allege, alleviate, allocate, allow,  allowed, ally, alter, amalgamate, amass, amaze, amble, amend, amortize, amount, amplify, analyze, anchor,  announce, answer, antagonize, anticipate, apologize, appeal, appear, appease, append, applaud, apply,  appoint, appraise, appreciate, approach, approve, are, argue, arise, arm, arouse, arrange, arrest, arrive,  articulate, ask, assassinate, assemble, assert, assess, assign, assimilate, assist, associate, assuage,  assume, assure, atone, attach, attack, attain, attempt, attend, attention, attest, attract, attribute,  auction, audit, audition, augment, authorize, autograph, average, avert, avoid, await, back, backfire, bail,  balance, balk, balloon, ban, band, banish, bank, bankroll, bankrupt, bar, bargain, base, bash, batter, battle,  bear, beat, become, bedevil, beef, begin, behave, belie, believe, bellow, belong, bend, benefit, bet, bid, bill,  bite, blackmail, blame, blast, bleed, blend, bless, blink, block, blow, blunder, blunt, blur, board, boast, bode,  bog, bolster, bomb, boom, boost, born, borrow, bother, bottle, bounce, bow, brace, brag, branch, brave, brazen,  breach, break, breathe, breed, brew, bribe, bridge, brief, bring, broadcast, broaden, browbeat, brush, buck,  buckle, budge, buffer, buffet, build, bumble, bump, buoy, burn, bury, butt, buttress, buy, buy-back, buzz,  bypass, calculate, call, calm, cancel, cap, capitalize, capture, care, careen, caricature, carry, cash, cast,  catapult, catch, cater, cause, caution, cease, cede, celebrate, cement, centralize, certify, challenge,  change, channel, characterize, charge, chart, chase, chat, chauffeur, cheat, check, cheer, chew, chill, chisel,  choke, choose, chop, churn, cinch, circulate, circumvent, cite, claim, clamp, clarify, clash, classify, clean,  cleanse, clear, click, climb, cling, clip, clobber, close, cloud, clutter, coach, coax, code, co-exist, cohere,  co-host, coincide, collaborate, collapse, collect, combat, combine, come, command, commemorate, commend, comment,  commercialize, commit, communicate, compare, compel, compensate, compete, compile, complain, complement,  complete, complicate, comply, compound, comprise, compromise, compute, computerize, con, conceal, concede,  conceive, concentrate, concern, conclude, condemn, condone, conduct, confer, confirm, confiscate, conform,  confront, confuse, congratulate, connect, connote, conquer, consent, conserve, consider, consist, console,  consolidate, constitute, constrain, construct, construe, consult, consume, contact, contain, contemplate,  contemporize, contend, content, continue, contract, contrast, contribute, control, convene, convert, convey,  convict, convince, cook, cool, cooperate, coordinate, cope, co-produce, copy, corner, corral, correct,  correspond, co-sponsor, cost, cough, count, countenance, counter, counteract, counterprogram, court, cover,  crack, craft, crank, crash, crawl, creak, create, credit, crest, criminalize, crimp, criticize, crop, cross,  crumble, crunch, crush, cry, cuff, curb, cure, curl, curry, curtail, cushion, cut, dabble, damage, damp, dampen,  dance, dare, dash, date, deal, debate, debunk, debut, deceive, decide, declare, decline, decrease, deduct,  default, defeat, defect, defend, defer, define, deflect, defraud, defuse, degenerate, delay, delegate,  deliberate, delight, deliver, demand, demilitarize, demobilize, democratize, demolish, demonstrate, denude,  deny, depart, depend, depict, deposit, depress, deprive, derail, deregulate, describe, desert, deserve, design,  designate, desist, destabilize, destroy, detect, deter, deteriorate, determine, detract, develop, devise,  devote, dial, dictate, die, differ, differentiate, dig, digest, dignify, dilute, diminish, dip, direct, disagree,  disappear, disappoint, disarm, disassemble, disassociate, disband, discard, discern, disclose, discomfit,  disconnect, discontinue, discount, discourage, discover, discredit, discuss, disdain, disengage, disguise,  dish, dislike, dismantle, dismember, dismiss, disobey, disparage, dispatch, dispel, dispense, displace, display,  dispose, disprove, dispute, disqualify, disregard, disrupt, dissipate, dissociate, dissolve, dissuade,  distance, distinguish, distort, distract, distribute, disturb, diverge, diversify, divert, divest, divide, do,  doctor, document, dole, dominate, don, donate, double, doubt, downsize, draft, drag, drain, drape, draw, dream,  dress, drift, drill, drink, drive, drop, drown, drum, dry, dump, duplicate, dwarf, earmark, earn, ease, eat,  eavesdrop, ebb, echo, eclipse, economize, edge, edit, educate, effect, elaborate, elect, eliminate, elongate,  emasculate, embark, embarrass, embellish, embrace, emerge, emote, empathize, emphasize, emphaticize, employ,  empty, emulate, enable, enact, encapsulate, encompass, encounter, encourage, end, endanger, endeavor, endorse, endure, enforce, engage, engineer, enhance, enjoin, enjoy, enlarge, enlist, enroll, ensue, ensure, entail,  enter, entice, entitle, entrench, entrust, envision, equal, equate, equip, eradicate, erase, erect, erode, erupt,  escalate, escape, establish, estimate, evade, evaluate, evaporate, even, evolve, exacerbate, exaggerate,  examine, exceed, except, exchange, exclude, excorciate, excuse, execute, exempt, exercise, exhaust, exhibit,  exist, exit, exorcise, expand, expect, expedite, expel, experience, expire, explain, explode, exploit, explore,  export, expose, express, expunge, extend, extinguish, extort, extract, extricate, fabricate, face, facilitate,  factor, fade, fail, fall, falsify, familiarize, fantasize, fare, farm, fashion, favor, fear, feature, feed, feel,  fend, ferret, ferry, fester, fetch, field, fight, figure, file, fill, finance, find, fine, fine-tune, finger,  finish, fire, firm, fit, fix, flash, flatten, flaunt, flay, flee, flinch, flip, float, flood, flounder, flourish,  flow, fluctuate, flush, fly, focus, fog, foil, fold, follow, fool, foot, force, forecast, foresee, forfeit, forge,  forget, forgive, forgo, form, formulate, foster, frame, franchise, free, freeze, freight, fret, frighten, frolic,  frustrate, fuck, fudge, fuel, fulfill, function, fund, funnel, furnish, further, gain, galvanize, gamble, garden,  garner, gasp, gather, gauge, gender, generalize, generate, get, give, glamorize, glaze, glean, glide, gloat, gloss,  glut, go, gon, gore, govern, grab, grace, grant, grapple, grasp, grimace, ground, group, grow, growth, guarantee,  guard, guess, guide, gut, guzzle, hack, halt, halve, hammer, hamper, hand, handicap, handle, hang, happen, harass,  harm, hasten, hate, haul, haunt, have, head, heal, hear, heat, hedge, heed, heighten, help, herald, hesitate, hide,  highlight, hinder, hinge, hint, hire, hit, hoe, hold, holler, homer, hone, honor, hope, host, house, hum, hurry, hurt,  identify, idle, ignite, ignore, illuminate, illustrate, imagine, impact, impair, impede, implement, implicate,  imply, import, impose, impound, impress, imprison, improve, impugn, incarcerate, inch, include, incorporate,  increase, increased, incur, indemnify, indicate, indict, induce, indulge, industrialize, infiltrate, inflame,  inflate, inflict, influence, inform, infringe, infuriate, infuse, ingest, ingratiate, inhibit, initiate,  inject, injure, innovate, insist, inspect, inspire, install, instill, institute, insulate, insure, integrate,  intend, intensify, interconnect, interest, interfere, interpret, intervene, interview, intimate, introduce,  invade, invent, invest, investigate, invite, invoke, involve, irk, iron, isolate, issue, itemize, jack,  jeopardize, join, joke, jolt, judge, juggle, jump, junk, justify, keen, keep, key, kick, kidnap, kill, kiss,  knock, know, kowtow, label, lack, lag, land, languish, lash, last, laugh, launch, launder, lay, lead, lean,  leap, leapfrog, learn, lease, leave, lecture, legislate, legitimize, lend, lengthen, lessen, let, level,  levy, liberalize, license, lie, lift, light, lighten, like, limit, line, linger, link, liquefy, liquidate,  list, listen, live, load, loan, lobby, locate, lock, lodge, log, look, loom, loose, loosen, loot, lose, love,  lower, lunch, lure, mail, maintain, make, man, manage, mandate
maneuver, manipulate, manufacture, march, mark, market, marry, marvel, massage, master, match, materialize, matter, mature, maul, maximize, mean, measure, mediate, meet, meld, melt, mention, merge, merit, mesh, migrate, militate, milk, mimic, mince, mind, minimize, mirror, misinterpret, misrepresent, miss, mistreat, mitigate, mix, moan, mobilize, model, moderate, modernize, modify, mollify, monitor, monopolize, mop, mortgage, motivate, mount, move, mow, mull, multiply, muse, muster, mute, nail, name, narrow, navigate, naysay, need, negotiate, net, network, nod, nominate, normalize, notch, note, notice, notify, nudge, nullify, nurture, obfuscate, object, obscure, observe, obtain, obviate, occasion, occupy, occur, offend, offer, offset, omit, ooze, open, operate, oppose, opt, order, organize, originate, oust, outfit, outflank, outfly, outlast, outline, outpace, outperform, outsell, outshine, out-smart, outstrip, out-trade, outweigh, overbid, overcome, overemphasize, overhaul, overlook, overpower, overpurchase, overreact, override, overrule, oversee, overstate, overthrow, overturn, overwhelm, owe, own, pace, pack, package, paint, pair, pale, palm, pan, panic, parachute, parallel, parcel, park, parry, part, partake, participate, pass, patronize, pave, pay, peak, pedal, peddle, peer, penalize, penetrate, perform, permit, perpetuate, persist, persuade, peruse, petition, phase, pick, piece, pile, pin, pinch, ping, pinpoint, pit, pitch, placate, place, plague, plan, plant, play, plea, plead, please, plot, plow, pluck, plug, plummet, plunge, plur, pocket, point, police, polish, poll, pollinate, pollute, ponder, pop, popularize, populate, portend, portray, pose, position, possess, post, postpone, pot, pounce, pound, pour, practice, pray, precede, preclude, predict, predispose, pre-empt, prefer, premiere, prepare, prepay, pre-register, presage, prescribe, present, preserve, press, pressure, pretend, pre-try, prevail, prevent, price, privatize, probe, proceed, process, proclaim, prod, produce, profile, profit, program, progress, prohibit, project, prolong, promise, promote, prompt, prop, propel, propose, prosper, protect, protest, prove, provide, provoke, prune, publicize, publish, pull, pummel, pump, punch, punish, purchase, pursue, push, put, puzzle, qualify, quantify, quarrel, quell, question, quiet, quit, quiz, quote, rage, raid, rain, raise, rally, ramp, range, rank, rat, ratify, rationalize, rattle, rave, reach, react, read, readmit, reaffirm, realign, realize, reap, reappraise, rearm, rearrange, reason, reassert, reassess, reassume, reassure, reauthorize, rebound, rebuild, rebut, recall, recapture, receive, reckon, reclaim, recognize, recommend, reconcile, reconnect, reconsider, reconstruct, record, recoup, recover, recraft, recreate, recycle, redden, redeem, redefine, redesign, redevelop, redial, rediscover, redo, redound, redraw, reduce, re-emerge, re-enter, reestablish, re-establish, re-evaluate, re-examine, refer, refinance, refine, reflect, refocus, refocuses, reform, refrain, refund, refurbish, refuse, refute, regain, regard, regenerate, register, regret, regroup, regulate, rehabilitate, reignite, reimburse, reimpose, rein, reinforce, reinstate, reinvent, reinvest, reinvigorate, reject, rejoin, rejuvenate, rekindle, relate, relaunch, relax, release, relieve, relinquish, relish, relocate, rely, remain, remake, remark, remedy, remember, remind, remove, renege, renegotiate, renew, renounce, renovate, rent, reopen, reorganize, repair, repatriate, repay, repeal, repeat, repel, replace, replaster, replenish, replicate, reply, repond, report, reposition, repossess, represent, reproduce, repurchase, request, require, reschedule, rescind, rescue, research, resell, resemble, resent, reserve, reset, reshape, reshuffle, reside, resign, resist, resolve, resort, respect, respond, rest, restart, restate, restore, restrain, restrict, restructure, result, resume, resurrect, retail, retain, rethink, retire, retreat, retrieve, retrofit, retry, return, reunite, revamp, reveal, reverberate, reverse, review, revise, revisit, revive, revoke, revolutionize, revolve, reward, rewrite, rid, ride, ring, rise, risk, rival, rock, roil, roll, roost, root, rotate, round, row, rub, rubber-stamp, ruin, rule, run, rush, sabotage, sacrifice, safeguard, safety, salvage, sap, satisfy, saturate, save, savor, say, scale, scan, scape, scare, schedule, school, scold, score, scorn, scour, scout, scrap, scream, screen, scrutinize, scurry, scuttle, seal, search, secure, seduce, see, seek, seem, segregate, seize, select, self-reinsure, sell, send, sense, sensitize, separate, sequester, serve, service, set, settle, sever, shadow, shake, shall, shape, share, sharpen, shave, shed, shell, shield, shift, shine, ship, shirk, shock, shoe-horn, shoot, shop, shore, shorn, short, shorten, shoulder, shout, shove, show, shower, shrink, shun, shut, shy, side, sidetrack, sift, sign, signal, signify, simplify, sing, sink, sit, ski, skid, skim, skip, skipper, slack, slam-dunk, slash, sleep, slide, slip, slog, slow, slump, smash, smell, smile, smoke, smooth, smother, smuggle, snatch, sniff, soak, soar, sob, socialize, sock, soften, solicit, solidify, solve, soothe, sort, sound, sour, sow, spare, spark, spawn, speak, specialize, specify, speculate, speed, spell, spend, spill, spin, split, sponsor, spot, spotlight, spread, spring, sprout, spruce, spur, spurn, spy, square, squeeze, stabilize, stack, staff, stage, stain, stall, stamp, stampede, stanch, stand, standardize, star, stare, start, starve, stash, state, staunch, stave, stay, steal, steer, stem, step, sterilize, stick, stifle, stimulate, stipulate, stir, stock, stockpile, stomach, stop, store, strafe, straighten, strain, stray, streamline, streetspeak, strengthen, stress, stretch, strike, strip, stroll, structure, struggle, study, stumble, subcontract, subject, sublet, submit, subordinate, subpoena, subscribe, subsidize, substantiate, substitute, subtract, subvert, succeed, suck, sue, suffer, suffice, suggest, suit, sum, summarize, summon, supersede, supervise, supplement, supply, support, suppose, suppress, surface, surge, surpass, surprise, surrender, surround, survey, survive, suspect, suspend, sustain, swallow, swamp, swap, sway, swear, sweat, sweeten, swell, swing, switch, synthesize, tackle, tag, take, talk, tamper, tandy, tap, target, tarnish, taste, tax, teach, team, tear, tell, tend, tender, term, terminate, test, test-drive, testify, thank, the, think, thrash, threaten, thrive, throw, thwart, tick, tie, tighten, tilt, time, tiptoe, tolerate, top, topple, torment, torpedo, torture, toss, total, totter, touch, tough, toughen, tour, tout, tower, trace, track, trade, traduce, trail, train, transact, transfer, transform, translate, transmit, transplant, transport, trash, travel, tread, treat, trend, trick, trickle, trigger, trim, triple, trivialize, trust, try, tumble, turn, twitch, unblock, uncover, undercut, undergo, underlie, underline, undermine, underperform, underpin, underscore, understand, undertake, underwrite, undo, undulate, unfold, unite, unleash, unload, unmask, unplug, unravel, unveil, unwind, update, upgrade, uphold, upset, urge, use, usurp, utilize, vacate, vacillate, vacuum, value, vanish, vary, vault, veer, vent, venture, verify, veto, view, violate, visit, visualize, vitiate, voice, void, volunteer, vote, wad, wade, wage, wail, wait, waive, wake, walk, wall, wan, wander, wane, want, ward, warm, warn, warrant, wash, waste, watch, water, weaken, wear, weather, wedge, weigh, weight, welcome, were, whack, whip, widen, will, wimp, win, wind, wipe, wire, wish, withdraw, withhold, withstand, witness, wonder, woo, work, worry, worsen, wound, wrap, wreak, wreck, wrest, wrestle, wring, write, yank, yield, zero, zip, zoom

The rest of the lemmas from \citet{giantVerbList} are available here: \href{https://patternbasedwriting.com/1/Giant-Verb-List-3250-Verbs.pdf}{https://patternbasedwriting.com/1/Giant-Verb-List-3250-Verbs.pdf}.

\end{document}